\documentclass{sigchi}

\toappear{\scriptsize Permission to make digital or hard copies of all or part of this work for personal or classroom use is granted without fee provided that copies are not made or distributed for profit or commercial advantage and that copies bear this notice and the full citation on the first page. Copyrights for components of this work owned by others than the author(s) must be honored. Abstracting with credit is permitted. To copy otherwise, or republish, to post on servers or to redistribute to lists, requires prior specific permission and/or a fee. Request permissions from permissions@acm.org. \\
{\emph{L@S '20, August 12--14, 2020, Virtual Event, USA.} } \\
\copyright~2020 Association for Computing Machinery. \\
ACM ISBN 978-1-4503-7951-9/20/08\ ...\$15.00.\\
DOI: http://dx.doi.org/10.1145/3386527.3405945}

\usepackage{balance}       
\usepackage{graphics}      
\usepackage[T1]{fontenc}   
\usepackage{txfonts}
\usepackage{mathptmx}
\usepackage[pdflang={en-US},pdftex]{hyperref}
\usepackage{color}
\usepackage{booktabs}
\usepackage{textcomp}

\usepackage{amsmath}
\usepackage{multirow}
\usepackage{comment}

\DeclareMathOperator{\argmax}{argmax}

\usepackage{microtype}        
\usepackage{ccicons}          

\usepackage{todonotes}

\def\plaintitle{Towards an Appropriate Query, Key, and Value\\Computation for Knowledge Tracing}

\def\plainkeywords{Education; Personalized learning; Knowledge Tracing; Deep Learning; Transformer}

\makeatletter
\def\url@leostyle{%
  \@ifundefined{selectfont}{
    \def\UrlFont{\sf}
  }{
    \def\UrlFont{\small\bf\ttfamily}
  }}
\makeatother
\def\pprw{8.5in}
\def\pprh{11in}

\definecolor{linkColor}{RGB}{6,125,233}
\hypersetup{%
  pdftitle={\plaintitle},
  pdfkeywords={\plainkeywords},
  pdfdisplaydoctitle=true, 
  bookmarksnumbered,
  pdfstartview={FitH},
  colorlinks,
  citecolor=black,
  filecolor=black,
  linkcolor=black,
  urlcolor=linkColor,
  breaklinks=true,
  hypertexnames=false
}

\begin{document}

\title{\plaintitle}

\numberofauthors{1}
\author{%
  \alignauthor{Youngduck Choi\textsuperscript{1,2},
Youngnam Lee\textsuperscript{1},
Junghyun Cho\textsuperscript{1},
Jineon Baek\textsuperscript{1,3},
Byungsoo Kim\textsuperscript{1}, \\
Yeongmin Cha\textsuperscript{1},
Dongmin Shin\textsuperscript{1},
Chan Bae\textsuperscript{1,4},
Jaewe Heo\textsuperscript{1}\\
    \affaddr{\textsuperscript{1}Riiid! AI Research, \textsuperscript{2}Yale University, \textsuperscript{3}University of Michigan, \textsuperscript{4}UC Berkeley}\\
    \email{\{youngduck.choi, yn.lee, jh.cho, jineon.baek, byungsoo.kim,\\
    ymcha, dm.shin, chan.bae, jwheo\}@riiid.co}}\\
}

\maketitle

\begin{abstract}
Knowledge tracing, the act of modeling a student's knowledge through learning activities, is an extensively studied problem in the field of computer-aided education.
Armed with attention mechanisms focusing on relevant information for target prediction, recurrent neural networks and Transformer-based knowledge tracing models have outperformed traditional approaches such as Bayesian knowledge tracing and collaborative filtering.
However, the attention mechanisms of current state-of-the-art knowledge tracing models share two limitations.
Firstly, the models fail to leverage deep self-attentive computations for knowledge tracing.
As a result, they fail to capture complex relations among exercises and responses over time.
Secondly, appropriate features for constructing queries, keys and values for the self-attention layer for knowledge tracing have not been extensively explored.
The usual practice of using exercises and interactions (exercise-response pairs), as queries and keys/values, respectively, lacks empirical support.

In this paper, we propose a novel Transformer-based model for knowledge tracing, SAINT: \textbf{S}eparated Self-\textbf{A}ttent\textbf{I}ve \textbf{N}eural Knowledge \textbf{T}racing.
SAINT has an encoder-decoder structure where the exercise and response embedding sequences separately enter, respectively, the encoder and the decoder.
The encoder applies self-attention layers to the sequence of exercise embeddings, and the decoder alternately applies self-attention layers and encoder-decoder attention layers to the sequence of response embeddings.
This separation of input allows us to stack attention layers multiple times, resulting in an improvement in area under receiver operating characteristic curve (AUC).
To the best of our knowledge, this is the first work to suggest an encoder-decoder model for knowledge tracing that applies deep self-attentive layers to exercises and responses separately.

We empirically evaluate SAINT on a large-scale knowledge tracing dataset, \emph{EdNet}, collected by an active mobile education application, \emph{Santa}, which has 627,347 users, 72,907,005 response data points as well as a set of 16,175 exercises gathered since 2016.
The results show that SAINT achieves state-of-the-art performance in knowledge tracing with an improvement of $1.8\%$ in AUC compared to the current state-of-the-art model.
\end{abstract}

\ccsdesc[500]{Social and professional topics~Student assessment}

\keywords{\plainkeywords}

\printccsdesc

\section{Introduction}
Creating a personalized educational agent that provides learning paths adapted to each student's ability and needs is a long-standing challenge of artificial intelligence in education.
Knowledge tracing, a fundamental problem for developing such an agent, is the task of predicting a student's understanding of a target subject based on their learning activities over time.
For example, one can predict the probability a student correctly answering a given exercise.
Tracing the state of a student's understanding enables efficient assignment of resources tailored to their ability and needs.

Traditional approaches to knowledge tracing include Bayesian knowledge tracing \cite{corbertt_1994} and collaborative filtering \cite{thai2010recommender, lee2016machine}.
With the advances in deep learning for applications to machine translation, healthcare and other modalities, neural network architectures such as Recurrent Neural Networks (RNNs) and Transformer \cite{vaswani_2017} 
have become the common building blocks in knowledge tracing models.
These models effectively capture the complex nature of students' learning activities over time, which is often represented as high-dimensional and sequential data.
In particular, \cite{lee_2019, liu_2019} use RNNs with attention to predict the probability distribution of a student's response to an exercise.
The model in \cite{pandy_2019} serves the same purpose, but it is based on the Transformer model which leverages a self-attention mechanism.
As a student goes through different exercises, their skills are correlated with the responses they gave to previous exercises.
Therefore, attention mechanisms are a natural choice for knowledge tracing because they learn to capture the inter-dependencies among exercises and responses, and give more weight to entries relevant for prediction \cite{luong2015effective, vaswani_2017}.

\begin{figure}[t]
\centering
\includegraphics[width=0.45\textwidth]{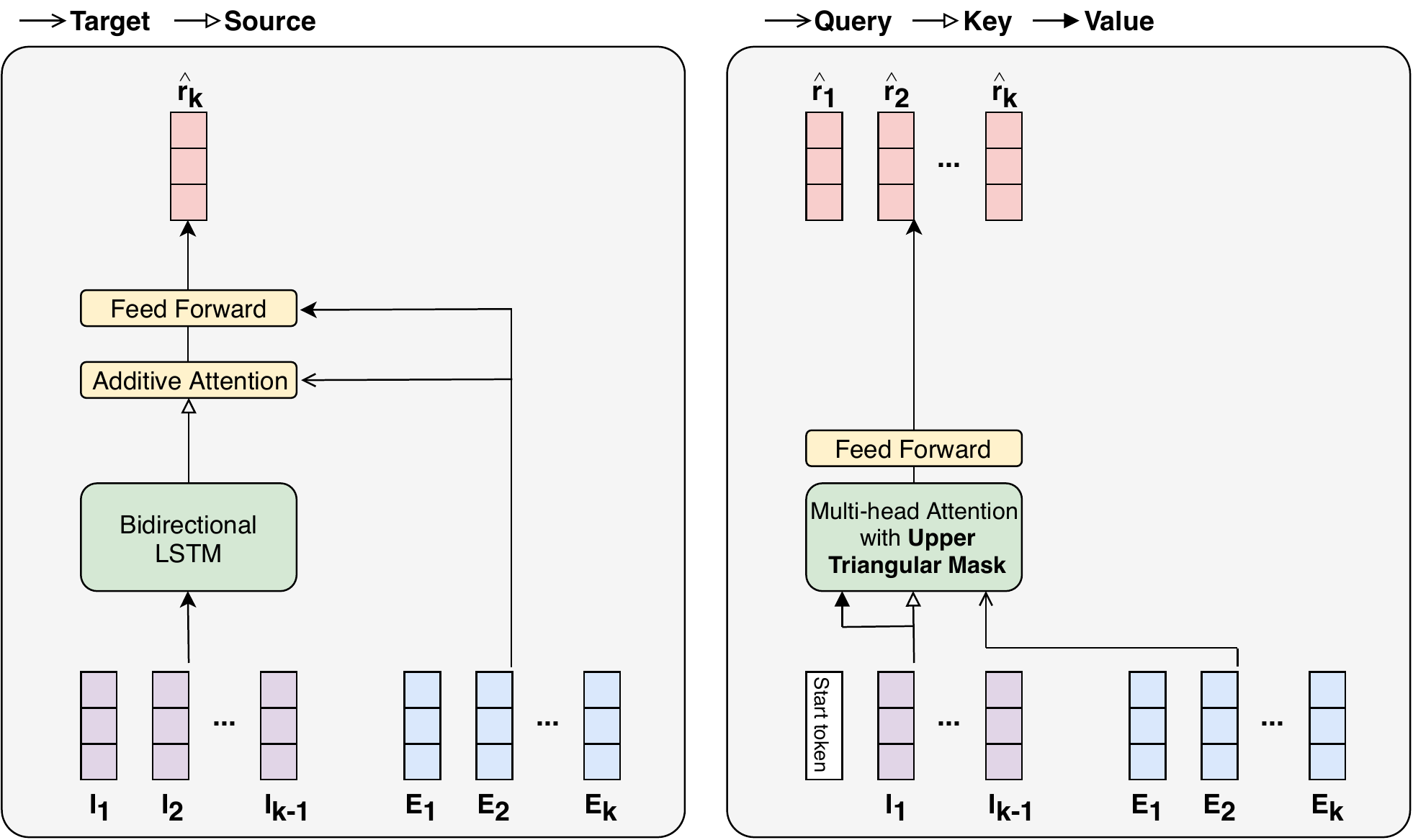}
\caption{Architectures of previous models. There are two limitations of the attention mechanisms used in  previous knowledge tracing models.
First, the models have attention layers too shallow to capture the complex relationships among different exercises and responses.
Second, the models rely on the same recipe: exercises for queries and interactions for keys and values.}
\label{fig:related_work}
\end{figure}

Despite their strengths, the attention mechanisms applied to knowledge tracing currently have two limitations as shown in Figure \ref{fig:related_work}.
First, previous models have attention layers too shallow to capture the possibly complex relations among different exercises and responses.
In particular, models in \cite{lee_2019,liu_2019,pandy_2019} have only one attention layer, and \cite{pandy_2019} shows a decrease in performance when self-attention layer is stacked multiple times.
Secondly, appropriate features for constructing queries, keys and values suited for knowledge tracing have not been explored thoroughly.
Given a series of interactions (exercise-response pairs) of a student, previous works rely on the same recipe: exercises for queries and interactions for keys and values \cite{lee_2019,liu_2019, pandy_2019}.
Other choices may provide a substantial gain in performance and thus need to be tested.

\begin{figure}[t]
\centering
\includegraphics[width=0.35\textwidth]{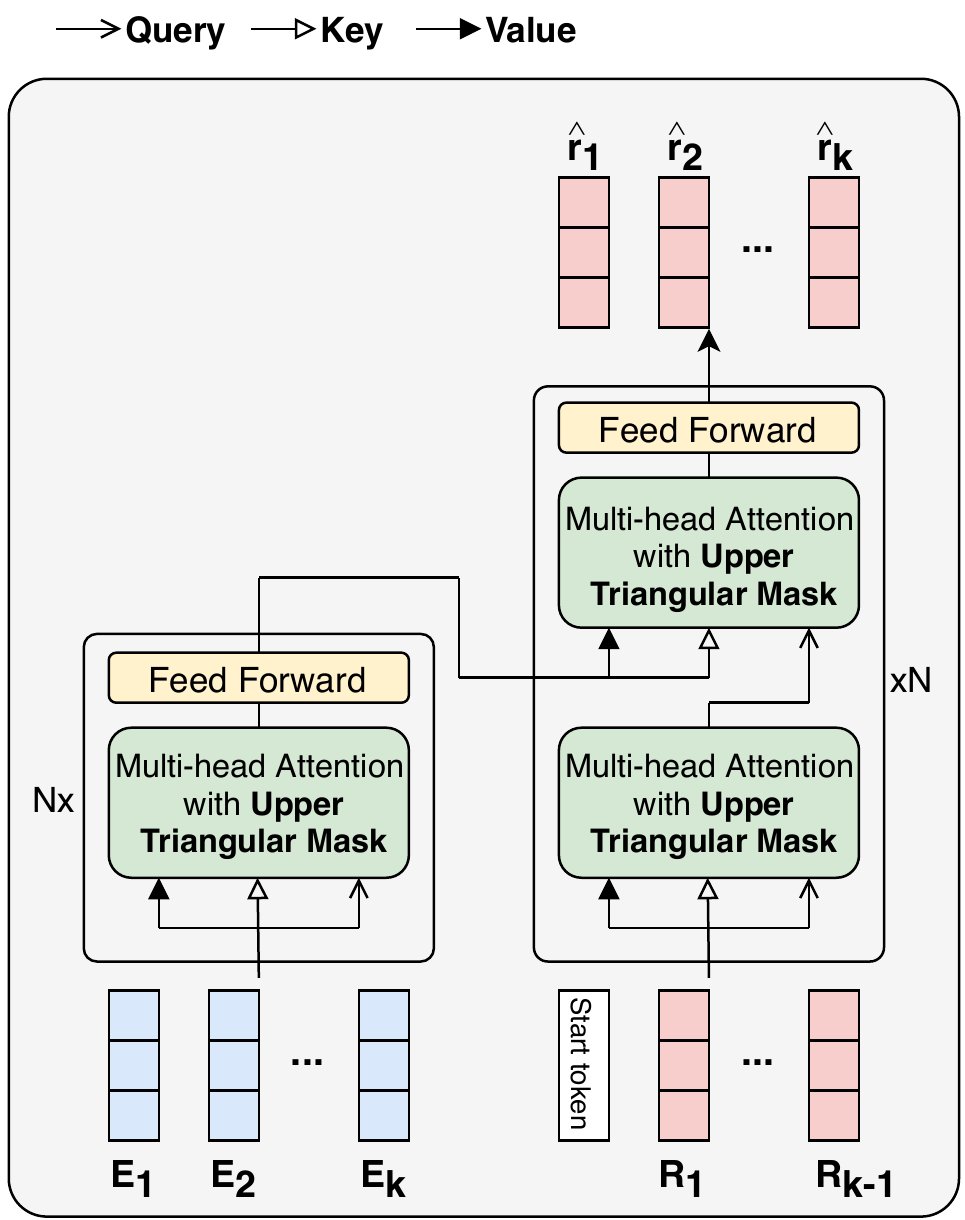}
\caption{The architecture of our proposed model, SAINT.
Unlike previous models, separating the exercise sequence and the response sequence and applying the encoder and the decoder, respectively, are the distinctive features that allow SAINT to capture complex relations among exercises and responses through deep self-attentive computations.}
\label{fig:architecture}
\end{figure}

In this paper, we address the problem of finding appropriate methods for constructing queries, keys and values for knowledge tracing.
Supported by a series of extensive empirical explorations, we propose a novel Transformer-based model for knowledge tracing, SAINT: \textbf{S}eparated Self-\textbf{A}ttent\textbf{I}ve \textbf{N}eural Knowledge \textbf{T}racing.
SAINT consists of an encoder and a decoder which are stacks of several identical layers composed of multi-head self-attention and point-wise feed-forward networks as shown in Figure \ref{fig:architecture}.
The encoder takes the sequence of exercise embeddings as queries, keys and values, and produces output through a repeated self-attention mechanism.
The decoder takes the sequential input of response embeddings as queries, keys and values then alternately applies self-attention and attention layers to the encoder output.
Compared to current state-of-the-art models, separating the exercise sequence and the response sequence, and feeding them to the encoder and decoder,respectively, are distinctive features that allow SAINT to capture complex relations among exercises and responses through deep self-attentive computations.

We conduct extensive experimental studies on a large scale knowledge tracing dataset, \emph{EdNet} \cite{choi2019ednet}, collected by an active mobile education application, \emph{Santa}, which has 627,347 users and 72,907,005 response data points on a set of 16,175 exercises gathered since 2016.
We compare SAINT with current state-of-the-art models and Transformer-based variants of deep knowledge tracing models.
Our experimental results show that SAINT outperforms all other competitors and achieves state-of-the-art performance in knowledge tracing as measured by area under receiver operating characteristic curve (AUC) with an improvement of $1.8\%$ compared to the current state-of-the-art model, SAKT \cite{pandy_2019}.

In summary, we make the following contributions:
\begin{itemize}
    \item We propose SAINT, a novel Transformer based encoder-decoder model for knowledge tracing where the exercise embedding sequence and the response embedding sequence separately enter the encoder and the decoder respectively.
    \item We show that SAINT effectively captures complex relations among exercises and responses using deep self-attentive computations.
    \item We empirically show that SAINT achieves a $1.8\%$ gain in AUC compared to the current state-of-the-art knowledge tracing model.
\end{itemize}

\section{Related Works}
Knowledge tracing is the modeling of a student's state of knowledge over time as they go through different learning activities.
It is an extensively studied problem in the field of computer-aided education.
For example, one can predict the probability of a student correctly answering a given exercise.
Such understanding enables efficient assignment of resources tailored to each student's ability and needs.

Traditional approaches to knowledge tracing include Bayesian Knowledge Tracing (BKT) \cite{corbertt_1994}, Collaborative Filtering (CF) \cite{thai2010recommender, lee2016machine} and many others.
BKT is a classical, prominent approach to knowledge tracing that has been studied extensively over time \cite{corbertt_1994, d2008more, yudelson2013individualized, pardos2010modeling, pardos2010navigating, pardos2013adapting, van2013properties, qiu2011does, kasurinen2009estimating, sao2013incorporating}.
BKT represents a student's knowledge state as a tuple of binary values.
Each value represents whether a student understands an individual concept or not.
The values are updated by a hidden Markov model using actual student responses. 

With the rise of Deep Learning (DL), DL-based knowledge tracing models such as Exercise-aware Knowledge Tracing (EKT) \cite{liu_2019}, Neural Pedagogical Agent (NPA) \cite{lee_2019} and Self-Attentive Knowledge Tracing (SAKT) \cite{pandy_2019} have been shown to outperform traditional models.
EKT and NPA are Bidirectional Long-Short Term Memory (Bi-LSTM) models with an attention mechanism built on the top layer.
SAKT is a Transformer based model with exercises as attention queries and past interactions (exercise-response pairs) as attention keys/values.
For effective knowledge tracing, one needs to analyze the relationships between the different parts of learning activity data.
The attention mechanism serves that purpose by weighing the more relevant parts of data heavier for prediction.

However, DL-based knowledge tracing models currently have two limitations.
First, the attention layer is shallow, so it is not able to capture the complex nature of students' learning activities over time which are often represented as high-dimensional and sequential data.
The networks in \cite{pandy_2019} do not apply self-attention to interactions and exercises, but use the latent features of the embedding layer directly as the input to the attention layer.
Also, they only use a single attention layer between interactions and exercises.
The networks in \cite{lee_2019, liu_2019} have deep LSTM layers that analyze the sequential nature of interactions, but the exercises are embedded directly and, again, they only use a single layer of attention between interactions and exercises.

Second, different combinations of features for attention queries, keys and values suited for knowledge tracing have not been thoroughly explored.
All of \cite{lee_2019, liu_2019, pandy_2019} use the same recipe: Given a series of interactions of a student, they use exercises to build the queries and interactions to build the keys/values.
Other possibilities, for instance, using self-attention on exercises or responses, to name one possibility among many, may provide substantial improvement and thus should be tested.

\section{Proposed Model}
\subsection{Problem Definition}
Given the history of how a student responded to a set of exercises, SAINT predicts the probability that the student will answer a particular new exercise correctly.
Formally, the student activity is recorded as a sequence $I_1, \cdots, I_n$ of interactions $I_i = (E_i, R_i)$.
Here $E_i$ denotes the \emph{exercise information}, the $i$-th exercise given to the student with related metadata such as the type of the exercise.
Similarly, the \emph{response information} $R_i$ denotes the student response $r_i$ to $E_i$ with related metadata such as the duration of time the student took to respond.
The student response $r_i \in{\{0,1\}}$ is equal to $1$ if their $i$-th response is correct and $0$ if it is not.
Thus, SAINT is designed to predict the probability 
\begin{align*}
    P(r_k = 1 | I_1, \cdots, I_{k-1}, E_k)
\end{align*}
of a student answering the $i$'th exercise correctly. 

\subsection{Input Representation}
\begin{figure}[t]
\centering
\includegraphics[width=0.3\textwidth]{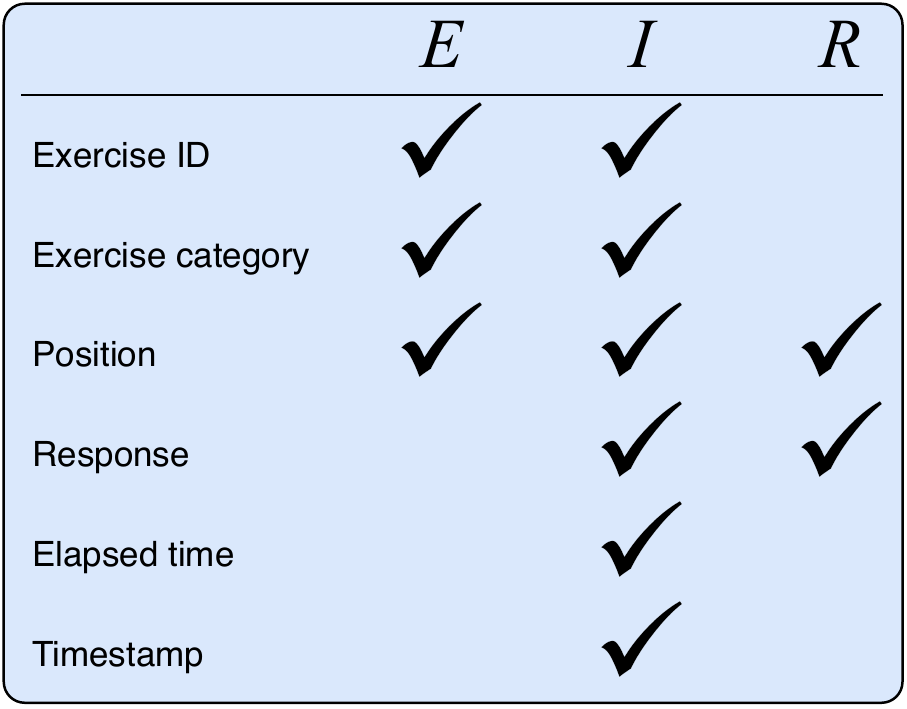}
\caption{We use a total of five different attributes to construct the input embeddings throughout the experiments. 
The exercise embedding is the sum of the exercise ID, exercise category and position embeddings. 
The response embedding is the sum of the response value and position embeddings.
Although interaction embeddings are not used in SAINT, Transformer-based variants of deep knowledge tracing models take them as an input.}
\label{fig:emb}
\end{figure}

SAINT takes the sequences of exercise information $E_1, \cdots, E_k$ and response information $R_1, \cdots, R_{k-1}$ as input, and predicts the $k$'th user response $r_k$.
The embedding layer in SAINT maps each $E_i$ and $R_i$ to a vector in latent space, producing a sequence of exercise embeddings $E_1^e, \cdots, E_k^e$ and a sequence of response embeddings $R_1^e, \cdots, R_{k-1}^e$.
The layer embeds the following attributes of $E_i$ and $R_i$.
\begin{itemize}
    \item Exercise ID: A latent vector is assigned to an ID unique to each exercise.
    \item Exercise category: Each exercise belongs to a category of the domain subject.
    A latent vector is assigned to each category.
    \item Position: The position (1st, 2nd, ...) of an exercise or a response in the input sequence is represented as a position embedding vector. 
    The position embeddings are shared across the exercise sequence and the response sequence.
    \item Response: A latent vector is assigned to each possible value (0 or 1) of a student's response $r_i$.
    \item Elapsed time: The time a student took to respond in seconds is rounded to an integer value.
    A latent vector is assigned to each integer between 0 and 300, inclusive. Any time more than 300 seconds is capped off to 300 seconds.
    \item Timestamp: Month, day and hour of the absolute time when a student received each exercise is recorded.
    A unique latent vector is assigned for every possible combination of month, day and hour.  
\end{itemize}
The final embedding $E_i^e$ or $R_i^e$ is the sum of all embedding vectors of its constituting attributes (see Figure \ref{fig:emb}). 
For example, the exercise information $E_i$ consists of the exercise ID, the exercise category, and the position, so we construct the exercise embedding $E_i^e$ by summing the corresponding exercise ID embedding, category embedding, and position embedding. 
Although we do not directly embed the interaction $I_i$ in SAINT, we will use interaction embeddings in Transformer-based variants of deep knowledge tracing models for comparison.

\subsection{Deep Self-Attentive Encoder-Decoder}
Our proposed model SAINT, based on the Transformer architecture proposed in \cite{vaswani_2017}, consists of an encoder and a decoder as shown in Figure \ref{fig:architecture}.
The encoder takes the sequence $E^e = [E_1^e, \cdots, E_k^e]$ of exercise embeddings and feeds the processed output $O = [O_1, \cdots, O_k]$ to the decoder.
The decoder takes $O$ and another sequential input $R^e = [S, R_1^e, \cdots, R_{k-1}^e]$ of response embeddings with the start token embedding $S$, and produces the predicted responses $\hat{r} = [\hat{r}_1, \cdots, \hat{r}_k]$.
\begin{align*}
O & = Encoder(E^e) \\
\hat{r} & = Decoder(O, R^e)
\end{align*}

The encoder and decoder are combinations of multi-head attention networks, which are the core components of SAINT, followed by feed-forward networks.
Unlike the original Transformer architecture, SAINT masks inputs corresponding to information from the future for all multi-head attention networks to prevent invalid attending.
This ensures that the computation of $\hat{r}_k$ depends only on the previous exercises $E_1, \cdots, E_k$ and responses $R_1, \cdots, R_{k-1}$.
We provide detailed explanations of the architecture in the following subsections.

\subsubsection{Multi-head Attention Networks} \hfill \\
The multi-head attention networks take $Q_\textrm{in}, K_\textrm{in}$ and $V_\textrm{in}$, each representing the sequence of queries, keys and values respectively.
The multi-head attention networks are simply the attention networks applied $h$ times to the same input sequence with different projection matrices.
An attention layer first projects each $Q_\textrm{in}, K_\textrm{in}$, and $V_\textrm{in}$ to a latent space by multiplying matrices $W_i^Q$, $W_i^K$ and $W_i^V$. That is,
\begin{align*}
Q_i = [q_1^i, \cdots, q_k^i] & = Q_\textrm{in} W_i^Q \\
K_i = [k_1^i, \cdots, k_k^i] & = K_\textrm{in} W_i^K \\
V_i = [v_1^i, \cdots, v_k^i] & = V_\textrm{in} W_i^V
\end{align*}
where $q$, $k$ and $v$ are the projected queries, keys and values, respectively.
The relevance of each value to a given query is determined by the dot-product between the query and the key corresponding to the value.

The attention networks in SAINT require a masking mechanism that prevents the current position from attending to subsequent positions.
The masking mechanism replaces upper triangular part of matrix $Q_i K_i^T$ from the dot-product with $-\infty$, which, after the softmax operation, has the effect of zeroing out the attention weights of the subsequent positions.
The attention head, $head_i$, is the values $V_i$ multiplied by the masked attention weights. That is,
$$head_i = \textrm{Softmax} \left( \textrm{Mask} \left(\frac{Q_i K_i^T}{\sqrt{d}} \right) \right) V_i$$ 
where the division by square root of $d$, the dimension of $q$ and $k$, is for scaling. 

A concatenation of $h$ attention heads is multiplied by $W^O$ to aggregate the outputs of different attention heads.
This concatenated tensor is the final output of the multi-head attention networks. 
$$\textrm{MultiHead}(Q_\textrm{in}, K_\textrm{in}, V_\textrm{in}) = \textrm{Concat}(head_1, \cdots, head_h) W^O$$

\subsubsection{Feed-Forward Networks} \hfill \\
Position-wise feed-forward networks are applied to the multi-head attention output to add non-linearity to the model,
\begin{align*}
F & = (F_1, \cdots, F_k) = \textrm{FFN}(M) \\
F_i & = \textrm{ReLU}\left(M_i W_1^{FF} + b_1^{FF} \right) W_2^{FF} + b_2^{FF}
\end{align*}
where $M = [M_1, \cdots, M_k] = \textrm{Multihead}(Q_\textrm{in}, K_\textrm{in}, V_\textrm{in})$ and $W_1^{FF}$, $W_2^{FF}$, $b_1^{FF}$ and $b_2^{FF}$ are weight matrices and bias vectors shared across different $M_i$'s.

\subsubsection{Encoder} \hfill \\
The encoder is a stack of $N$ identical layers which are the feed-forward networks followed by the multi-head attention networks.
In formula, each identical layer is
\begin{align*}
M & = \textrm{SkipConct}(\textrm{Multihead}(\textrm{LayerNorm}(Q_\textrm{in}, K_\textrm{in}, V_\textrm{in}))) \\
O & = \textrm{SkipConct}(\textrm{FFN}(\textrm{LayerNorm}(M)))
\end{align*}
where skip connection \cite{he2016deep} and layer normalization \cite{ba2016layer} are applied to each sub-layer.
Note that $Q_\textrm{in}$, $K_\textrm{in}$ and $V_\textrm{in}$ of the first layer are $E^e$, the sequence of exercise embeddings, and those of subsequent layers are the output of the previous layer.

\subsubsection{Decoder} \hfill \\
The decoder is also a stack of $N$ identical layers which consist of the multi-head attention networks followed by the feed-forward networks.
Similar to the encoder, skip connection and layer normalization are applied to each sub-layer.

Each identical layer is represented by following equations:
\begin{align*}
M_1 & = \textrm{SkipConct}(\textrm{Multihead}(\textrm{LayerNorm}(Q_\textrm{in}, K_\textrm{in}, V_\textrm{in}))) \\
M_2 & = \textrm{SkipConct}(\textrm{Multihead}(\textrm{LayerNorm}(M_1, O, O))) \\
L & = \textrm{SkipConct}(\textrm{FFN}(\textrm{LayerNorm}(M_2)))
\end{align*}
where $O$ is the final output of the encoder.
The $Q_\textrm{in}$, $K_\textrm{in}$ and $V_\textrm{in}$ of the first layer in the decoder are all $R^e$, the sequence of response with start token embeddings, and those of the following layers are the output of the previous layer.

Finally, a prediction layer, consisting of a linear transformation layer followed by a sigmoid operation, is applied to the output of last layer so that the decoder output is a series of probability values.

\begin{figure*}[t]
\centering
\includegraphics[width=0.9\textwidth]{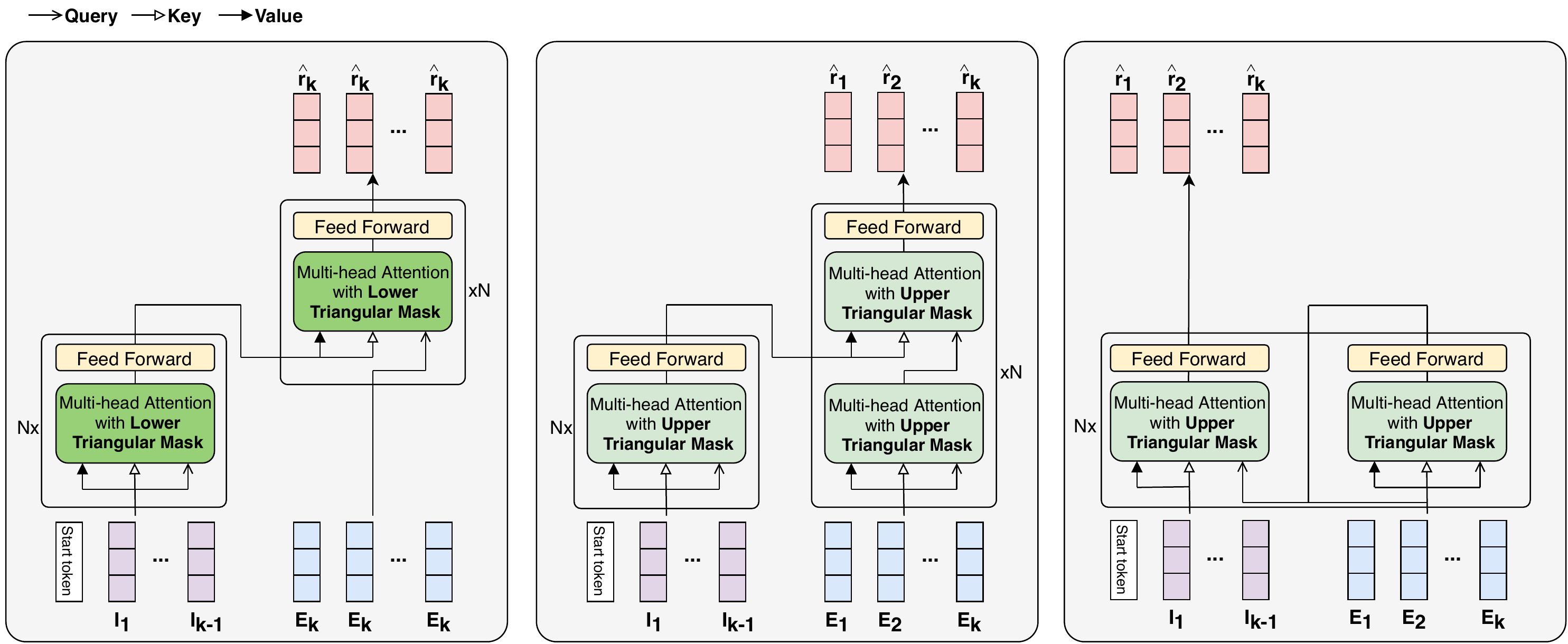}
\caption{We compare SAINT with three different architectures based on multiple stacked attention layers: Lower Triangular Masked Transformer with Interaction sequence (LTMTI), Upper Triangular Masked Transformer with Interaction sequence (UTMTI) and Stacked variant of SAKT (SSAKT) (from left to right).}
\label{fig:c234}
\end{figure*}

\subsection{Transformer based Variants of Deep Knowledge Tracing Models} \label{section:variant}
In order to find a deep attention mechanism effective for knowledge tracing, we conduct extensive experiments on different ways to construct the query, key and value for attention mechanisms.
Through the experiments, we explore a total of four different architectures with multiple stacked attention layers including SAINT.
In this section, we describe the architectures of the other three models: Lower Triangular Masked Transformer with Interaction sequence (LTMTI), Upper Triangular Masked Transformer with Interaction sequence (UTMTI) and Stacked variant of SAKT \cite{pandy_2019} (SSAKT) (see Figure \ref{fig:c234}).

\subsubsection{LTMTI} \hfill \\
The LTMTI model predicts the response $r_k$ by taking the exercise $E_k$ and the sequence of $k-1$ past interactions $I = [I_1, \cdots, I_{k-1}]$ separately.
Firstly, the encoder applies self-attention $N$ times to the sequence of past interactions $I$ plus a start token, producing the output $O = [O_1, \cdots, O_k]$.
Then, the first layer of the decoder takes the encoder output $O$ as keys and values, and the constant sequence $E_k, \cdots, E_k$ of size $k$ as queries.
The next $N-1$ layers of the decoder takes the encoder output $O$ as keys and values, and the output of the previous layer as queries.

A major computational difference of LTMTI from SAINT is that LTMTI applies lower triangular masks to all attention layers, instead of the upper triangular masks in SAINT.
This forces the $i$'th output $\hat{r}_{k, i}$ of LTMTI to be inferred only from $E_k$ and the most recent $(i-1)$ interactions. That is,
\begin{align*}
    \hat{r}_{k, 1} & = \argmax_{r} P(r_k = r | E_k) \\
    \hat{r}_{k, 2} & = \argmax_{r} P(r_k = r | I_{k-1}, E_k) \\
    & \vdots \\
    \hat{r}_{k, k} & = \argmax_{r} P(r_k = r | I_{1}, \cdots, I_{k-1}, E_k).
\end{align*}
Note that this aspect of LTMTI is in effect an augmentation on the training data where histories are truncated to various lengths.
That is, the model learns to infer $r_k$ with multiple past interaction histories $(E_k)$, $(I_{k-1}, E_k)$, $\cdots$, $(I_1, \cdots, I_{k-1}, E_k)$ from a single input.
For testing, the prediction $\hat{r}_k$ is defined to be the output $\hat{r}_{k, i}$ that uses the longest interaction sequence available.

\subsubsection{UTMTI} \hfill \\
The UTMTI model follows the same architecture as SAINT and differs only on the choice of input sequence.
Unlike SAINT, UTMTI takes the interaction sequence $I_1, \cdots, I_{k-1}$ as the encoder input and the exercise sequence $E_1, \cdots, E_k$ as the decoder input.
This method follows the attention mechanism of SAKT \cite{pandy_2019} and EKT \cite{liu_2019}. 
In SAKT, exercises are embedded as queries and interactions are embedded as keys and values. 
Likewise, EKT processes the interaction sequence with a deep Bi-LSTM layer, and combines the output with attention weights given by the cosine similarity of exercise embeddings.
UTMTI follows the same pattern of using interactions as attention keys/values and exercises as attention queries.

\subsubsection{SSAKT} \hfill \\
The SSAKT model consists of a single attention block that uses exercise embeddings as queries and interaction embeddings as keys/values.
The authors report a decrease in AUC when the attention block is stacked multiple times. 
SSAKT resolves this issue by applying self-attention on exercises before supplying them as queries.
The outputs of the exercise self-attention block and the exercise-interaction attention block enters the corresponding following blocks as inputs for their attention layers.

\begin{figure*}[t]
\centering
\includegraphics[width=0.9\textwidth]{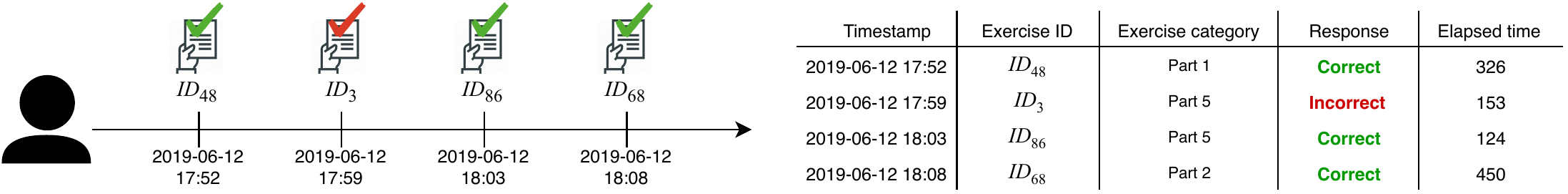}
\caption{Description of \emph{EdNet} dataset. For each user, the dataset contains his interaction history, which is a series of records each consisting of the following features: 
the absolute time when the user received each exercise (Timestamp), an ID unique to each exercise (Exercise ID), a category of the domain subject that each exercise belongs to (Exercise category), user response value (Response) and the time the user took to respond in seconds (Elapsed time).}
\label{fig:data}
\end{figure*}

\begin{figure}[t]
\centering
\includegraphics[width=0.5\textwidth]{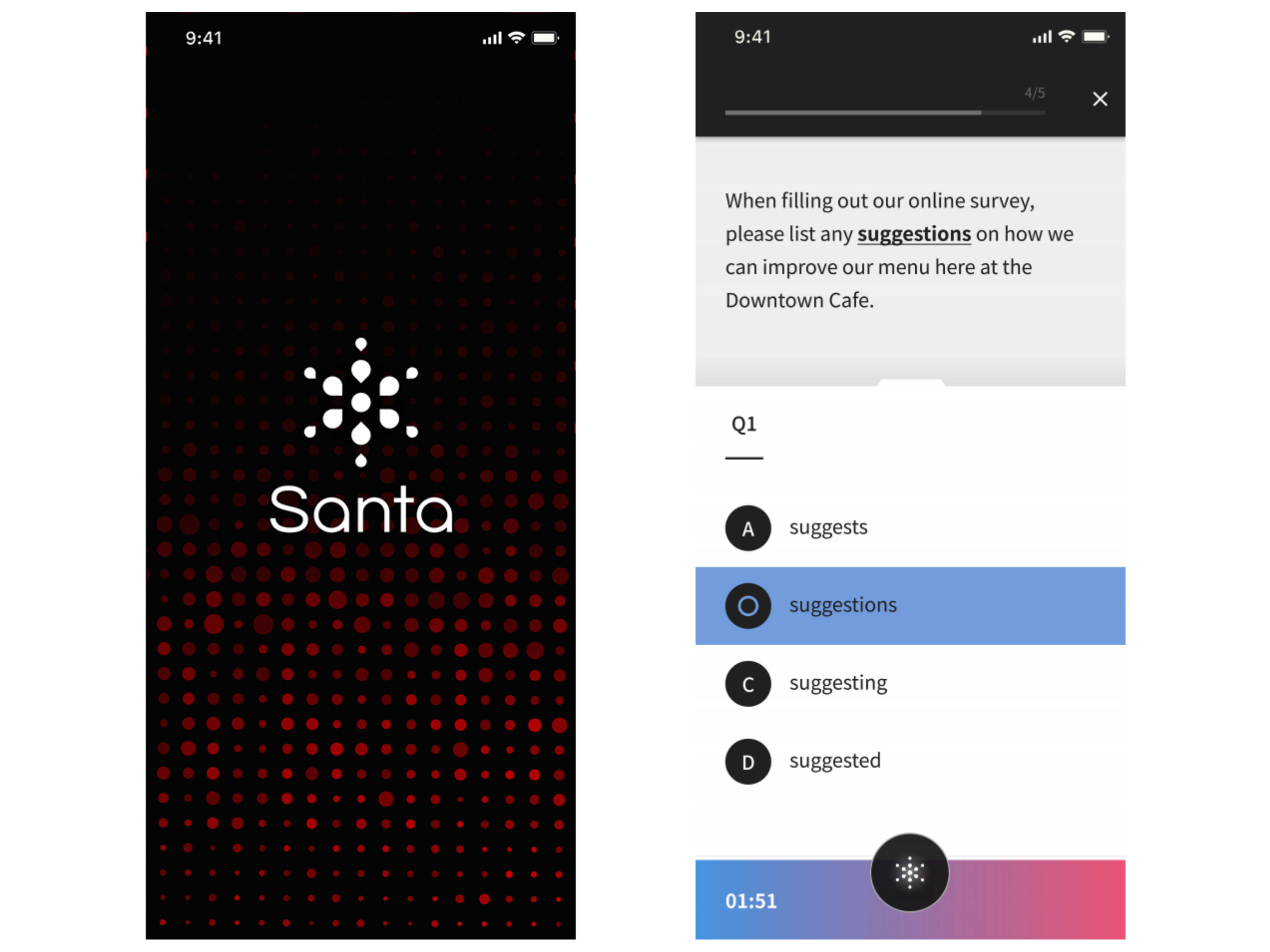}
\caption{User interface of \emph{Santa}.
\emph{Santa} is a self-study solution equipped with artificial intelligence tutoring system that aids students to prepare the Test of English for International Communication (TOEIC) Listening and Reading Test.}
\label{fig:ui}
\end{figure}

\section{Experiments}
\subsection{Dataset}
We conduct experiments on a large scale knowledge tracing dataset, \emph{EdNet} \cite{choi2019ednet}, collected by \emph{Santa}, an active mobile application for English education (Figure \ref{fig:ui}).
\emph{Santa} is a self-study solution equipped with an artificial intelligence tutoring system that helps students prepare the Test of English for International Communication (TOEIC) Listening and Reading Test.
\emph{Santa} currently has 1,047,747 registered users and is available on both Android and iOS.
The dataset consists of multiple-choice exercises with corresponding user responses that have been collected since 2016.
The dataset contains the interaction histories of each user.
An interaction history is a series of records, each consisting of the following features: the time at which the user received each exercise (Timestamp), an ID unique to each exercise (Exercise ID), a category of the domain subject that each exercise belongs to (Exercise category), the user response (Response), and the length of time taken by the user to respond (Elapsed time) (Figure \ref{fig:data}).
The dataset has a total of 16,175 exercises and 72,907,005 responses, with 627,347 users solving more than one exercise.
We split the dataset into three parts per user basis: the train set (439,143 users, 51,117,443 responses), the validation set (62,734 users, 7,460,486 responses) and the test set (125,470 users, 14,329,076 responses) (Table \ref{table:statistics}).

\begin{table}[t]
    \centering
    \begin{tabular}{c|c}
    \hline
    \toprule
     Statistics & \emph{EdNet} dataset \\
    \toprule
     total user count                   & 627,347 \\
     train user count                   & 439,143 \\
     validation user count              & 62,734 \\
     test user count                    & 125,470 \\
     total response count               & 72,907,005 \\
     train response count               & 51,117,443 \\
     validation response count          & 7,460,486 \\
     test response count                & 14,329,076 \\
     correct response ratio             & 0.66 \\
     incorrect response ratio           & 0.34 \\
     mean interaction sequence length   & 116.21 \\
     median interaction sequence length & 13 \\
     max interaction sequence length    & 41,644 \\
    number of categories                &7 \\
    \bottomrule
    \end{tabular}
    \caption{Statistics of \emph{EdNet} dataset}
    \label{table:statistics}
\end{table}

\begin{table}[h]
\centering
\begin{tabular}{ccc}
\hline
\toprule 
Methods   & ACC  & AUC \\
\toprule
MLP                     & 0.7052 & 0.7363   \\ 
NCF                   & 0.7051 & 0.7341   \\ 
\toprule
NPA                     & 0.7290 & 0.7656   \\ 
SAKT                    & 0.7271 & 0.7671   \\ 
\toprule
LTMTI & 0.7339 & 0.7762\\ 
UTMTI & 0.7323 & 0.7735\\ 
SSAKT & 0.7340 & 0.7777\\ 
\toprule
SAINT & \textbf{0.7368}  & \textbf{0.7811}\\ 
\bottomrule
\end{tabular}
\caption{Comparison of the current state-of-the-art models and SAINT}
\label{table:comparison}
\end{table}

\subsection{Training Details}\label{subsec:detail}
Hyper-parameters are determined by the ablation study in Section \ref{subsec:ablstudy}.
We train the model from scratch, using the Xavier uniform \cite{xavier_unifom} distribution to initialize weights. 
The window size, dropout rate, and batch size are set to 100, 0.1, and 128 respectively.
We use the Adam optimizer \cite{kingma2014adam} with $lr=0.001, \beta_1=0.9, \beta_2=0.999$ and $epsilon=1e-8$. 
We use the so-called Noam scheme to schedule the learning rate as in \cite{vaswani_2017} with $warmup\_steps$ set to 4000.
We pick the model parameters that give the best results on the validation set and evaluate them with the test set.

\begin{figure*}[t]
\centering
\includegraphics[width=1.0\textwidth]{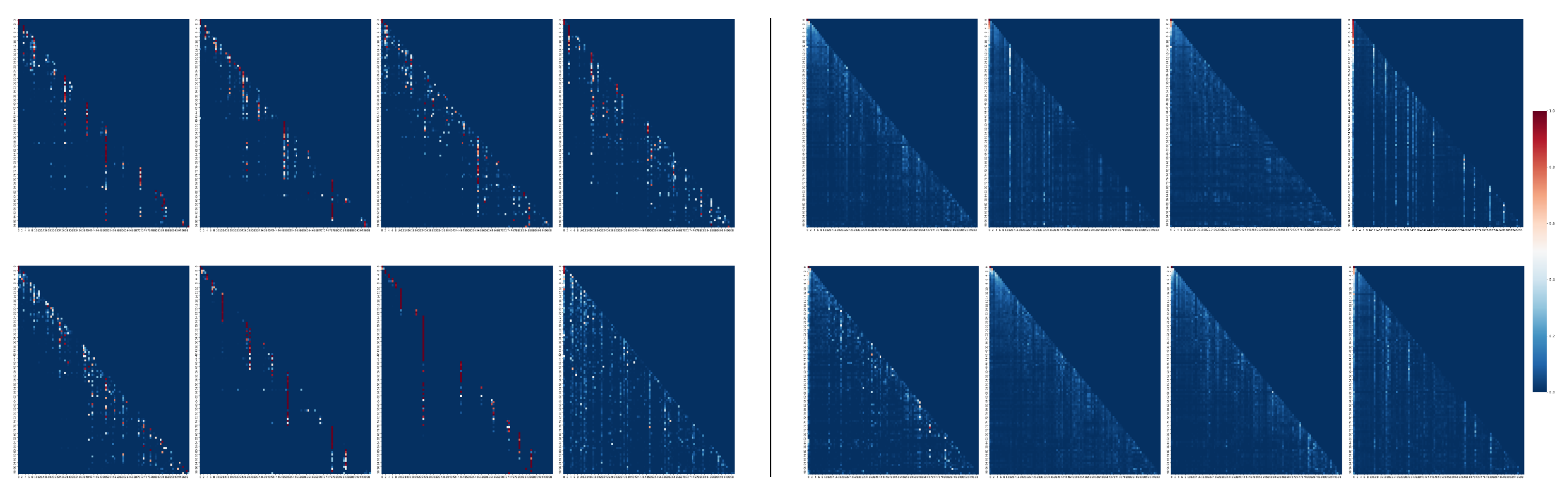}
\caption{Self-attention of last encoder block (left) and last decoder (right) block.
Each figure shows the attention matrix of each head in the self-attention layer. 
The attention matrices of the encoder are sparse - most exercises attend only a few relevant exercises.
The attention matrices of the decoder are dense - the attention values are spread over a large number of responses.}
\label{fig:sa}
\end{figure*}

\begin{figure}[t]
\centering
\includegraphics[width=0.5\textwidth]{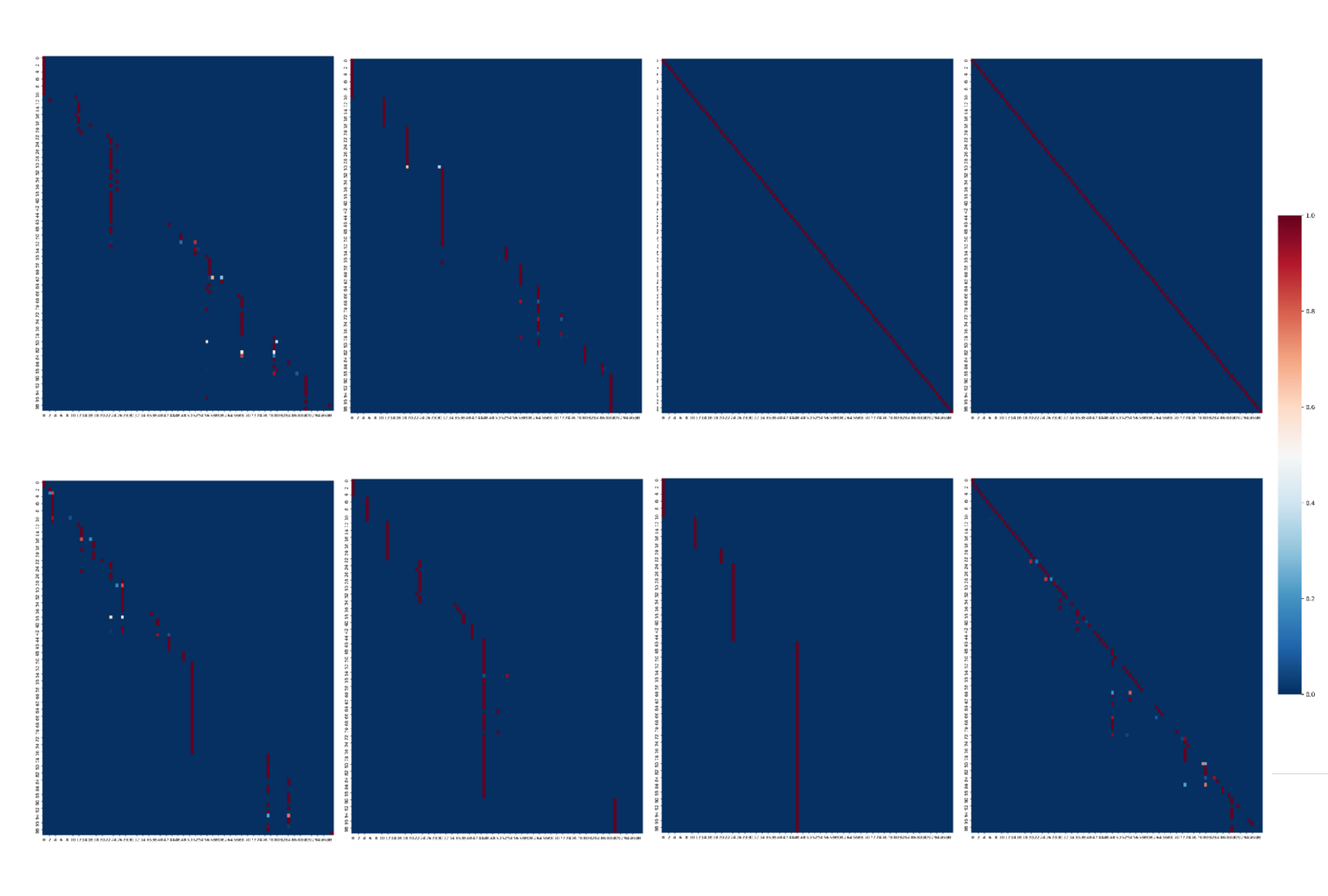}
\caption{Visualization of the encoder-decoder attention values of the first decoder block.
Some heads only have large weights on the diagonal, showing that each response attends only the corresponding exercise.
Other heads show a series of vertical stripes, showing that only a few exercises are attended to by all responses.}
\label{fig:en_de_mu}
\end{figure}

\begin{figure}[t]
\centering
\includegraphics[width=0.5\textwidth]{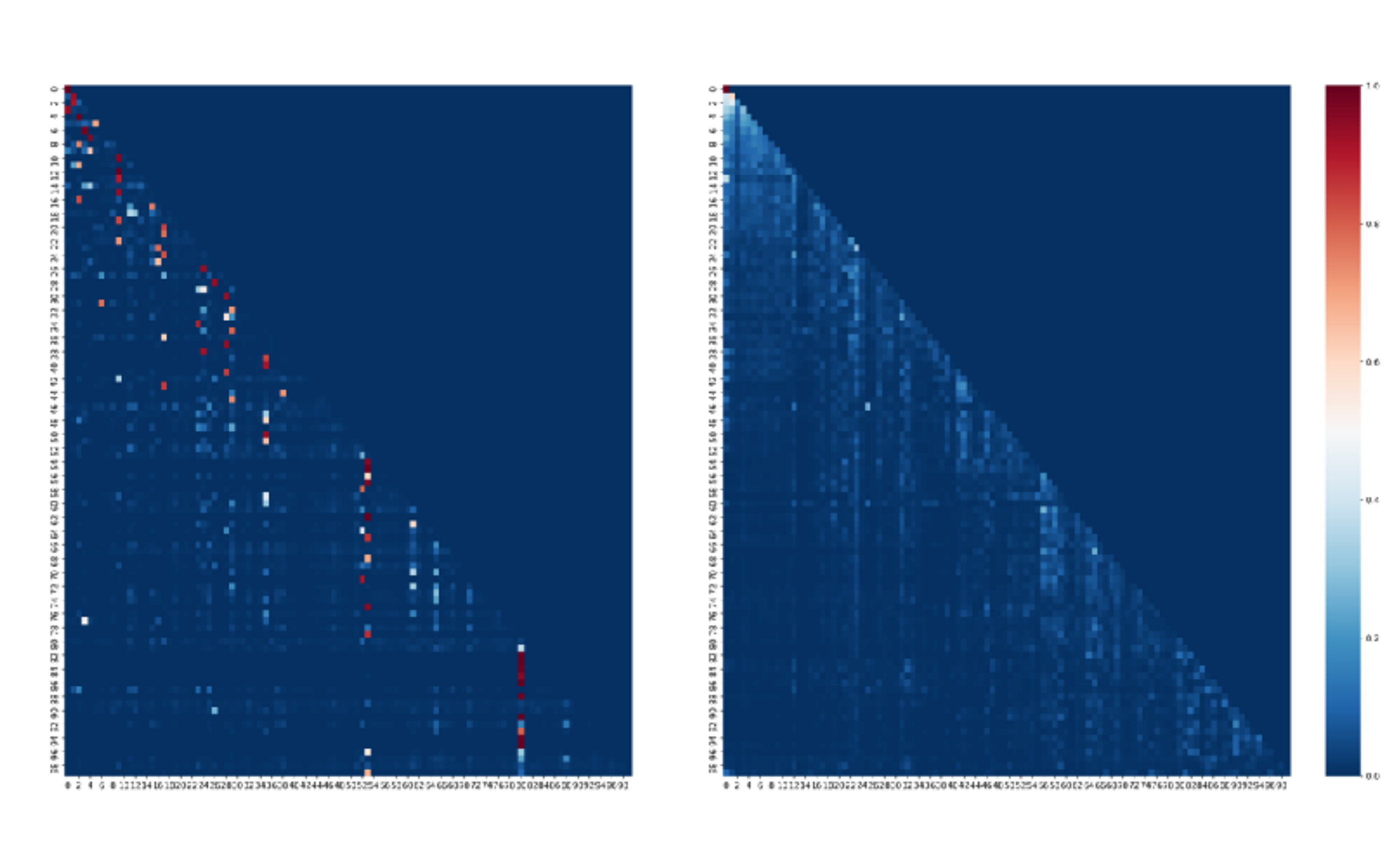}
\caption{Comparison of decoder self-attention matrices at varying depths. The attention values in the first decoder block (left) are sparse, showing that each query only attends a few relevant response values.
However, the attention values in the last decoder block (right) are more evenly distributed, showing that each query attends to the responses more thoroughly.}
\label{fig:att_lay}
\end{figure}

\begin{figure*}[t]
\centering
\includegraphics[width=0.74\textwidth]{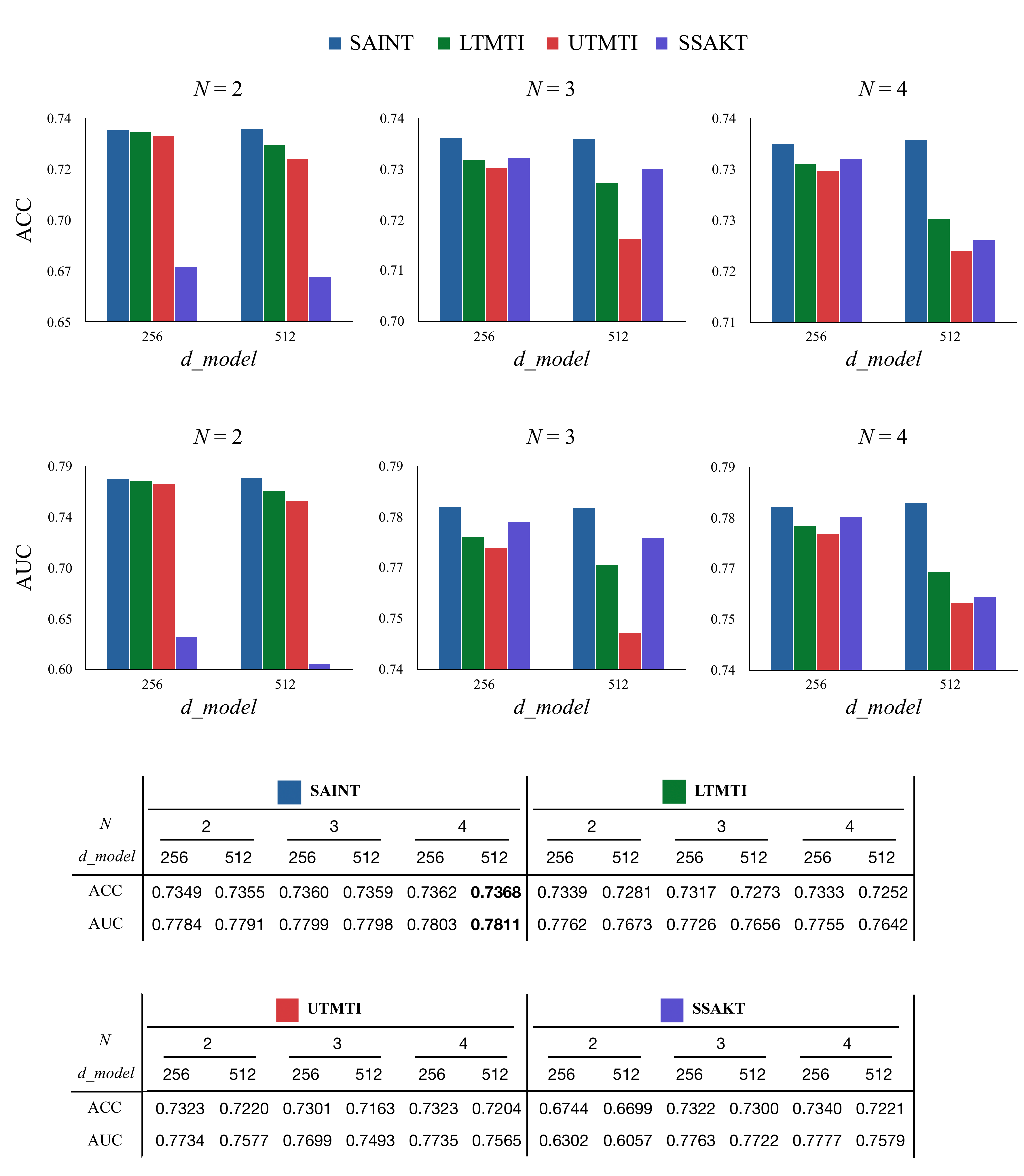}
\caption{Graph (above) and table (below) showing the ACC and AUC of our proposed methods at different model sizes.
$N$ is the number of stacked encoder and decoder blocks and $d\_model$ is the dimension of all sub-layer output of the model.}
\label{fig:ablation}
\end{figure*}

\begin{table}[t]
\centering
\begin{tabular}{cccccc}
\hline
\toprule 
\multirow{2}{*}{$N$}& \multirow{2}{*}{$d\_model$}& \multicolumn{2}{c}{Embedding A} & \multicolumn{2}{c}{Embedding B} \\ \cline{3-6} 
& & ACC  & AUC     & ACC    & AUC   \\
\toprule 
\multirow{2}{*}{2}  & 256         & 0.7349  & 0.7784  & 0.7340     & 0.7763   \\ 
                    & 512        & 0.7355  & 0.7791  & 0.7308     & 0.7715   \\ 
\toprule
\multirow{2}{*}{3}  & 256         & 0.7360  & 0.7799  & 0.7345     & 0.7773   \\ 
                    & 512        & 0.7359  & 0.7798  & 0.7309     & 0.7715   \\ 
\toprule
\multirow{2}{*}{4}  & 256         & 0.7362  & 0.7803  & 0.7348     & 0.7778   \\ 
                    & 512        & \textbf{0.7368}  & \textbf{0.7811}  & 0.7318     & 0.7741   \\ 
\bottomrule
\end{tabular}
\caption{Ablation study for embedding}
\label{table:embedding}
\end{table}

\subsection{Experimental Results}
We evaluate SAINT by comparing it with the current state-of-the-art knowledge tracing based approaches and collaborative filtering based approaches: Multilayer Perceptron (MLP) \cite{he2017neural}, Neural Collaborative Filtering (NCF) \cite{he2017neural}, Neural Pedagogical Agent (NPA) \cite{lee_2019} and Self-Attentive Knowledge Tracing (SAKT) \cite{pandy_2019}.
In our experiments, we use two performance metrics commonly used in previous works: the area under the receiver operating characteristic curve (AUC) and accuracy (ACC).
AUC shows sensitivity (recall) against $1-\text{specificity}$.
Sensitivity (resp. specificity) is the proportion of true positives (resp. negatives) that are correctly predicted to be positive (resp. negative). 
ACC is the proportion of predictions that were correct.
Table \ref{table:comparison} presents the overall results of our evaluation. It shows that our model outperforms the other models in both metrics.
Compared to the current state-of-the-art models, the ACC of SAINT is higher by 1.1\% and AUC is higher by 1.8\%. 
Other Transformer-based variants of deep knowledge tracing models described in Section \ref{section:variant} also perform better than existing approaches.

We visualize the attention weights of the fully trained, best-performing SAINT model to analyze the attention mechanism of SAINT.
The self-attention weights of the last encoder and decoder block show different tendencies (see Figure \ref{fig:sa}).
This shows that SAINT is capable of learning the different attention mechanisms appropriate for exercises and responses, and applying them separately.
Figure \ref{fig:en_de_mu} shows that each head of the first decoder block's encoder-decoder attention layer attends the exercise sequence in varying patterns. 
As seen in Figure \ref{fig:att_lay}, the span of attention in later decoder blocks are more diverse.
This can be interpreted as the attention mechanism capturing the increase in the complexity of the values that incorporate more complex relationships between the exercises and the responses as the values go through successive decoder blocks.

\subsection{Ablation Study}
\label{subsec:ablstudy}
In this section, we present ablation studies for the suggested models.
Firstly, we run ablation studies on each architecture with different hyper-parameters. 
Figure \ref{fig:ablation} shows that SAINT, which applies deep attention layers separately to exercises and responses, gives the best result.

The best performing model of SAINT has 4 layers and a latent space dimension of 512.
It shows an ACC of 0.7368 and AUC of 0.7811.
Next to SAINT, LTMTI models show high ACC and AUC overall.
This shows that the data augmentation effects from lower-triangular masks in LTMTI boosts ACC and AUC.

Secondly, we evaluate the best-performing SAINT model when trained with inputs of different levels of detail.
All models share the same embedding method for exercises. 
For response embeddings, Embedding A only uses positional information and the user response value to build the response embedding while Embedding B uses the exercise category, timestamp, and elapsed time in addition to the features used by Embedding A.
Table \ref{table:embedding} shows that using more information to construct embeddings did not improve results.

\section{Conclusion}
In this paper, we proposed SAINT, a state-of-the-art knowledge tracing model with deep attention networks.
We empirically demonstrated a Transformer-based architecture that shows superior performance on knowledge tracing tasks.
In addition, we showed through extensive experiments on the queries, keys, and values of attention networks that separately feeding exercises and responses to the encoder and decoder, respectively, is ideal for knowledge tracing tasks.
Furthermore, by investigating the attention weights of SAINT, we found that the results of self-attention of encoder and decoder exhibit different patterns. 
We suggested that this supports our idea that the separation of exercises and responses in input allows the model to find attention mechanisms that are especially suited to the respective input values.
Finally, Evaluation of SAINT on a large-scale knowledge tracing dataset showed that the model outperforms existing state-of-the-art knowledge tracing models.

\bibliographystyle{SIGCHI-Reference-Format}
\bibliography{ref}

\end{document}